\documentclass[10pt,twocolumn,letterpaper]{article}

\usepackage[pagenumbers]{cvpr} 
\usepackage{amsmath,amssymb}

\usepackage{colortbl}
\usepackage{booktabs}
\usepackage{multirow}
\usepackage{multicol}
\usepackage[mathscr]{euscript}
\usepackage[dvipsnames]{xcolor}
\usepackage{colortbl}
\usepackage{makecell}
\usepackage{subcaption}
\usepackage[symbol]{footmisc}

\usepackage{tcolorbox}
\usepackage{amssymb}
\usepackage{fontawesome5}

\usepackage[utf8]{inputenc} 
\usepackage[T1]{fontenc}    
\usepackage{url}            
\usepackage{booktabs}       
\usepackage{amsfonts}       
\usepackage{nicefrac}       
\usepackage{graphicx}
\usepackage{microtype}      
\usepackage{xcolor}         
\usepackage[pagebackref,breaklinks,colorlinks,citecolor=cvprblue]{hyperref}
\usepackage[capitalize]{cleveref}

\definecolor{cvprblue}{rgb}{0.21,0.49,0.74}
\usepackage[pagebackref,breaklinks,colorlinks,citecolor=cvprblue]{hyperref}
\def\paperID{10}
\def\confName{CVPRW}
\def\confYear{2025}

\title{Caption Anything in Video: Fine-grained Object-centric Captioning via Spatiotemporal Multimodal Prompting}

\author{%
  Yunlong Tang\textsuperscript{\rm 1}, Jing Bi\textsuperscript{\rm 1}, Chao Huang\textsuperscript{\rm 1}, Susan Liang\textsuperscript{\rm 1}, Daiki Shimada\textsuperscript{\rm 2}, Hang Hua\textsuperscript{\rm 1}, \\Yunzhong Xiao\textsuperscript{\rm 3}, Yizhi Song\textsuperscript{\rm 4}, Pinxin Liu\textsuperscript{\rm 1}, Mingqian Feng\textsuperscript{\rm 1}, Junjia Guo\textsuperscript{\rm 1}, Zhuo Liu\textsuperscript{\rm 1},\\ Luchuan Song\textsuperscript{\rm 1}, Ali Vosoughi\textsuperscript{\rm 1}, Jinxi He\textsuperscript{\rm 1}, Liu He\textsuperscript{\rm 4}, Zeliang Zhang\textsuperscript{\rm 1}, Jiebo Luo\textsuperscript{\rm 1}, Chenliang Xu\textsuperscript{\rm 1} \\
  \\
  \textsuperscript{\rm 1}University of Rochester, \textsuperscript{\rm 2}Sony Group Corporation, \textsuperscript{\rm 3}CMU, \textsuperscript{\rm 4}Purdue University
}

\begin{document}

\maketitle

\begin{abstract}
We present CAT-V (Caption AnyThing in Video), a training-free framework for fine-grained object-centric video captioning that enables detailed descriptions of user-selected objects through time.
CAT-V integrates three key components: a Segmenter based on SAMURAI for precise object segmentation across frames, a Temporal Analyzer powered by TRACE-Uni for accurate event boundary detection and temporal analysis, and a Captioner using InternVL-2.5 for generating detailed object-centric descriptions.
Through spatiotemporal visual prompts and chain-of-thought reasoning, our framework generates detailed, temporally-aware descriptions of objects' attributes, actions, statuses, interactions, and environmental contexts without requiring additional training data.
CAT-V supports flexible user interactions through various visual prompts (points, bounding boxes, and irregular regions) and maintains temporal sensitivity by tracking object states and interactions across different time segments.
Our approach addresses limitations of existing video captioning methods, which either produce overly abstract descriptions or lack object-level precision, enabling fine-grained, object-specific descriptions while maintaining temporal coherence and spatial accuracy.
The GitHub repository for this project is available at: \url{https://github.com/yunlong10/CAT-V}
\end{abstract}

\begin{figure}[h]
  \centering
  \includegraphics[width=\linewidth]{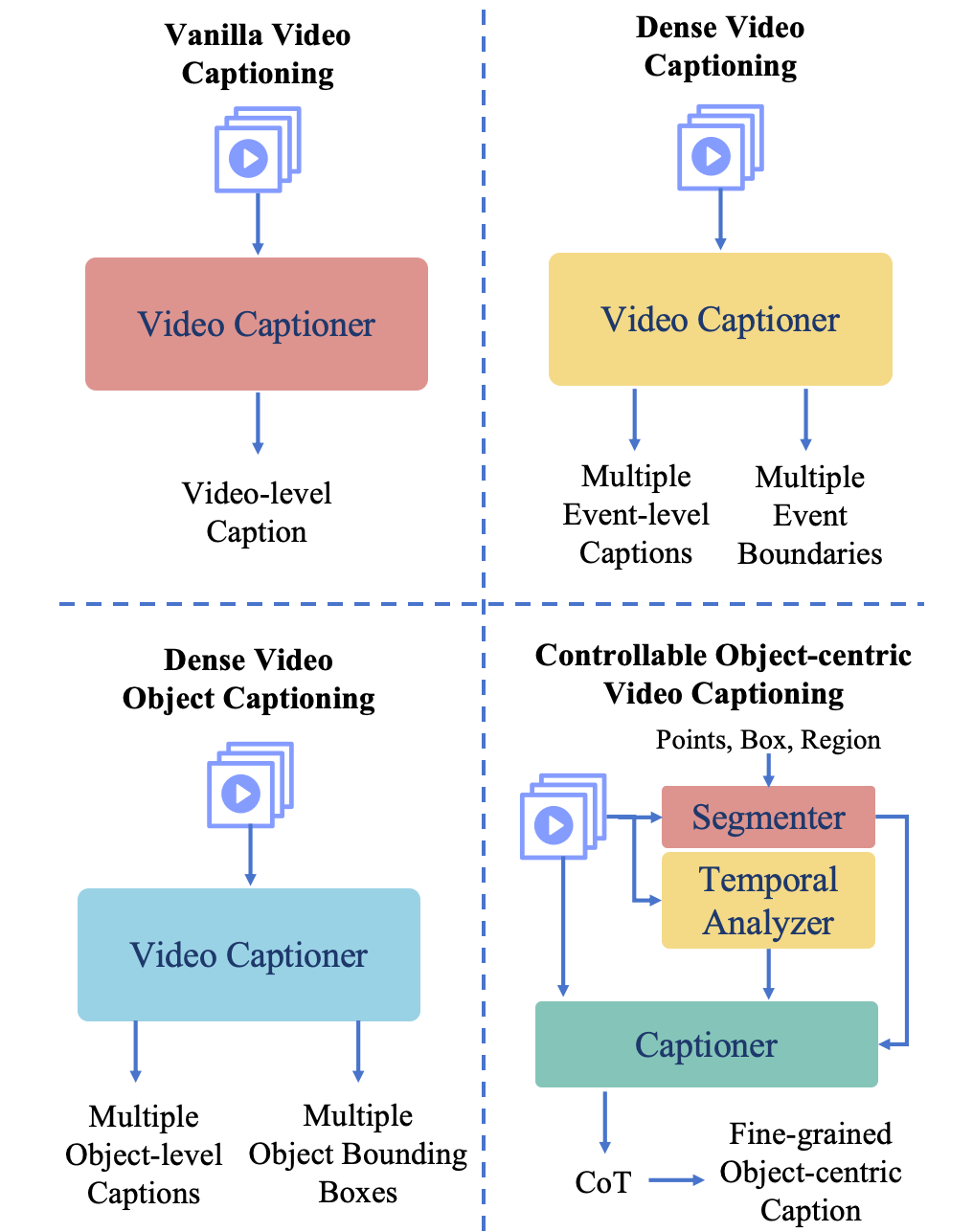}
  \caption{Comparison of video captioning approaches: Vanilla (top-left), Dense (top-right), Dense Object (bottom-left), and our CAT-V framework (bottom-right) with integrated modules for user-controlled object-centric captioning via integrated modules (Segmenter, Temporal Analyzer, Captioner with CoT reasoning).}
  \label{fig:teaser}
\vspace{-1em}
\end{figure}

\section{Introduction}

Video captioning, which aims to generate coherent natural language descriptions of video content, remains a fundamental challenge in vision-language learning.
Given a video as input, current multimodal large language models (MLLMs) that can handle video understanding tasks or video large language models (VidLLMs)~\cite{tang2023video} can be prompted to perform detailed vanilla video captioning, which is video-level and attempts to cover all aspects of the video content.
However, vanilla video captioning lacks the sensitivity and dynamics of time and space.
For instance, video is dynamic~\cite{Huang_2023_CVPR,zhang2021unicon}, and the same object can perform various actions at different times during the video, but most of the existing VidLLMs for general purposes~\cite{maaz2023videochatgpt,li2023videochat} tend to generate too abstract answers, which are more suitable for captioning static images.
Dense video captioning (DVC) involves generating multiple captions for multiple events along with their temporal boundaries.
However, the current task-specific model designed for DVC~\cite{wang2021pdvc} tends to produce excessively concise outputs. Some existing works explore VidLLMs-based methods~\cite{yang2023vid2seq,tang2024avicuna,huang2024vtimellm,guo2024trace,zhang2025videollama3} that are fine-tuned on dense video captioning datasets~\cite{krishna2017dense,zhou2018towards}, but they somewhat compromise the ability to follow instructions and still struggle with more fine-grained, object-centric captioning.
These methods also lack effective user interaction and only provide a language interface for users.
While some works have investigated controllable image captioning~\cite{wang2023caption,huang2024segment}, controllable fine-grained object-centric captioning in videos remains underexplored. Additionally, some studies~\cite{yuan2025sa2va} have sought to integrate the Segment Anything Model (SAM) with MLLMs/VidLLMs; however, these methods depend on annotated data for training both MLLMs and SAM.

To address these limitations, we introduce \emph{\textbf{C}aption \textbf{A}ny\textbf{T}hing in \textbf{V}ideo} (CAT-V), a training-free framework for \emph{object-centric video captioning} augmented by a pre-trained segmentation model built on VidLLMs.
CAT-V consists of three main components: a Segmenter, a Temporal Analyzer, and a Captioner.
\Cref{fig:teaser} illustrates the key differences between our proposed approach and existing video captioning methods, highlighting how CAT-V integrates user control, object-level focus, and temporal awareness in a unified framework.
Specifically, the Segmenter is a pre-trained video object segmentation model based on an improved version of SAM 2~\cite{ravi2024sam}, known as SAMURAI~\cite{yang2024samurai}. It generates pixel-level masklets of an object throughout the entire video as indicated by the user within a single frame of the input video.
Benefiting from the training of SAM 2, CAT-V supports a range of visual prompts, including points and bounding boxes, to accurately identify the object desired by the user during interactions.
The original video is then updated by injecting the predicted masklets of the selected object, which serve as spatiotemporal visual prompts.
The Temporal Analyzer is based on TRACE-Uni~\cite{guo2024trace}, a temporal-aware VidLLM pre-trained on dense video captioning datasets, enabling CAT-V to perceive the events and changes occurring in the video, produce coarse-grained event-level captions, and identify the corresponding boundaries.
The Captioner is based on InternVL-2.5~\cite{chen2024internvl2.5} and takes the spatiotemporal prompted updated video as input, along with the temporal boundaries and coarse-grained event captions provided by the Temporal Analyzer. 
The Captioner also accepts Chain-of-Thought (CoT) prompting as input.
This approach encourages the Captioner to focus on the object selected/highlighted by the user, sufficiently identifying the object's attributes, actions, and statuses, the environments or backgrounds surrounding the object, any other objects interacting with the selected object, and events related to the selected object, ultimately generating fine-grained object-centric captions.

Different from previous controllable captioning methods~\cite{yuan2025sa2va}, CAT-V is training-free and does not rely on a large amount of annotated data for training or fine-tuning, sufficiently utilizing the capabilities of pre-trained MLLMs/VidLLMs. 
Besides, CAT-V provides an efficient interaction mode for users to select the object that they want to accurately and fine-grained describe in the video, well inherent in the flexibility of SAM 2, where the limitation of previous general VidLLMs~\cite{maaz2023videochatgpt,li2023videochat}, which could not interact through visual prompts, has been lifted.
Moreover, by utilizing the temporal awareness of Trace-Uni, CAT-V is sensitive to dynamic changes in events related to the selected object, making it possible to capture the status changes.
We present these strong capabilities of CAT-V through a comprehensive array of qualitative examples in the experimental results.

\noindent In short, our contribution is twofold:
\begin{itemize}
    \item We propose CAT-V, a training-free framework for object-centric video captioning that leverages pre-trained models to generate fine-grained descriptions without requiring additional training data, addressing the limitations of existing video captioning approaches.
    \item We demonstrate that CAT-V achieves temporal-aware and spatially-precise object-centric video captioning by combining the temporal analysis capabilities of TRACE-Uni with the spatial segmentation abilities of SAMURAI, enabling detailed descriptions of object.
\end{itemize}

\begin{figure*}[!ht]
  \centering
  \includegraphics[width=1\linewidth]{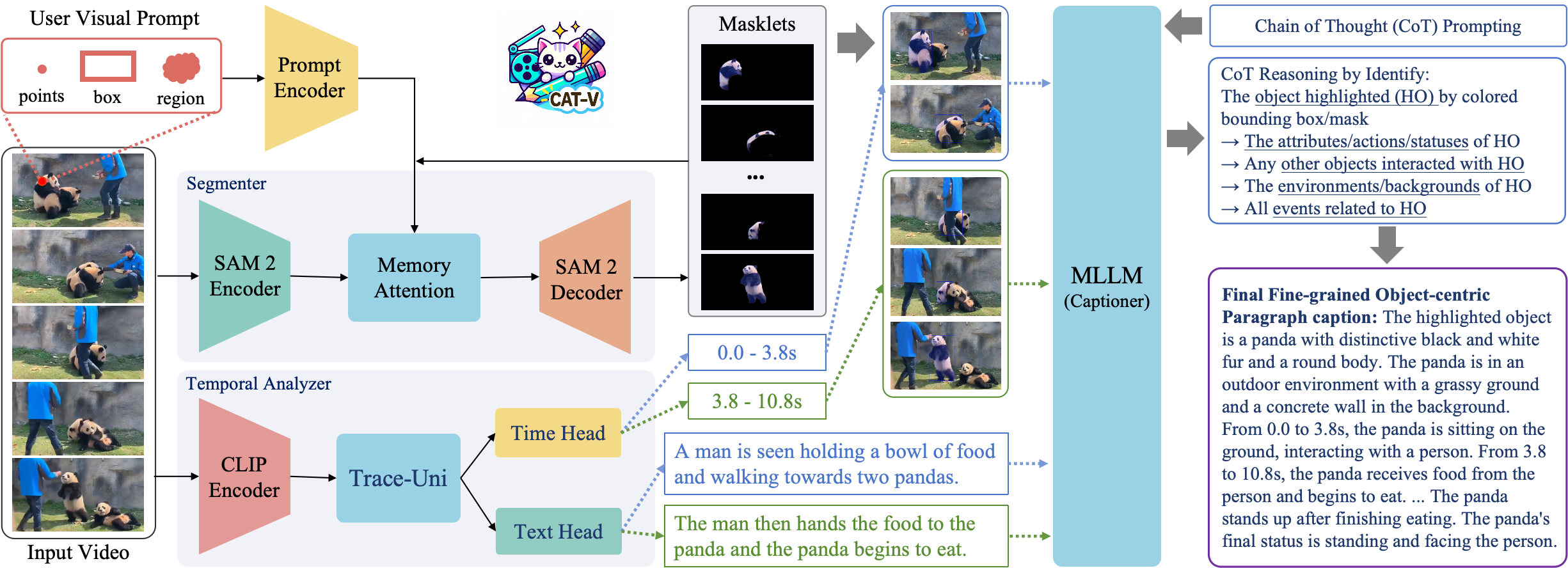}
  \caption{CAT-V consists of three modules: Segmenter, Temporal Analyzer, and Captioner. The Segmenter precisely segments objects in video frames using user-defined prompts (points, bounding boxes, or regions). The Temporal Analyzer captures video dynamics hierarchically. The Captioner creates object-centric captions using upstream information and CoT reasoning.}
  \label{fig:model}
\end{figure*}

\section{CAT-V: Caption Anything in Video}

Our proposed framework, CAT-V, is designed for fine-grained object-centric video captioning via spatiotemporal multimodal prompting. 
It integrates three key modules: the Segmenter $\mathcal{S}$, the Temporal Analyzer $\mathcal{T}$, and the Captioner $\mathcal{C}$. 
This modular approach allows for dynamic user visual input, points or bounding boxes, and irregular regions, to guide the generation of detailed and contextually relevant captions.
\Cref{fig:model} illustrates the architecture of CAT-V.
Given an input video $V = \{I_t\}_{t=1}^T$ with $T$ frames and a user prompt $p$, the framework operates as follows.

\subsection{Segmenter}
The Segmenter $\mathcal{S}$, powered by SAMURAI~\citep{yang2024samurai}, performs precise object segmentation in video frames based on user-provided visual prompts.
For each frame $I_t$, the Segmenter produces a binary mask $M_t = \mathcal{S}(I_t, p)$ where $M_t \in \{0,1\}^{H \times W}$ represents the pixel-level segmentation of the target object, with $H$ and $W$ being the frame height and width respectively.
The module uses the SAM 2's encoder~\cite{ravi2024sam} to embed the input video frames, a prompt encoder to encode the user visual prompt, and SAM 2's decoder.
SAMURAI enhances the capabilities of SAM 2 with Kalman filtering and motion-aware memory, enabling robust object mask extraction even in challenging scenarios with occlusions, motion blur, or complex backgrounds.

\subsection{Temporal Analyzer}
The Temporal Analyzer $\mathcal{T}$, built upon TRACE-Uni~\citep{guo2024trace}, models the temporal dynamics of video sequences through a hierarchical approach.
It processes the video $V$ to identify $N$ events with their corresponding temporal boundaries $\{(s_i, e_i)\}_{i=1}^N$, where $s_i$ and $e_i$ represent the start and end timestamps of the $i$-th event. 
For each event, it generates a coarse-grained caption $c_i = \mathcal{T}(V, s_i, e_i)$. 
This temporal decomposition enables fine-grained analysis of object interactions and activities across different time scales.

\subsection{Captioner}
The Captioner $\mathcal{C}$, an MLLM implemented using InternVL-2.5-8B~\citep{chen2024internvl}, generates detailed object-centric captions by integrating multiple inputs: the original video $V$, object masks $\{M_t\}_{t=1}^T$, temporal event boundaries $\{(s_i, e_i)\}_{i=1}^N$, coarse-grained event captions $\{c_i\}_{i=1}^N$, and chain-of-thought prompts $P_{CoT}$. 
This ensures that the generated captions are both spatially precise and temporally coherent.
The final object-centric caption is generated as:
$$C_{final} = \mathcal{C}(V(\{M_t\}_{t=1}^T, f), \{(s_i, e_i, c_i)\}_{i=1}^N, P_{CoT}),$$
where $f$ controls how the masklets are injected into the original video (introduced in \Cref{sec:injection}).

\subsection{Chain-of-Thought Prompting}
We design fine-grained prompts to guide the Captioner in Chain-of-Thought (CoT) reasoning, enabling systematic and structured analysis of object-centric video content.
Our prompting strategy can be represented as a sequence of analytical components $P_{CoT} = \{A_1, A_2, ..., A_K\}$, where each component $A_k$ focuses on a specific aspect of object analysis (attributes, actions, status changes, etc.). 
This structured approach helps the model first identify and analyze individual aspects before synthesizing them into a coherent, temporally-aware narrative.
By explicitly separating these analytical components, we ensure that no critical details are overlooked in the final description.

\tcbset{
  colback=green!3!white,
  colframe=green!30!black,
  boxrule=0.5pt,
  arc=6pt,
  left=4pt,
  right=4pt,
  top=5pt,
  bottom=5pt,
  boxsep=4pt,
}
\begin{tcolorbox}[]
{\Large\textcolor{yellow!70!orange}{\faLightbulb}} \textbf{Chain-of-Thought Prompting}\\
Above are the event captions given by the user, whose timestamps are very accurate but the subjects of the sentences are not necessarily what we want to highlight. Please pay attention to the object highlighted (HO) by colored bounding box and blue mask in the video frames, and generate accurate object-centric caption for the HO. Please make sure in object-centric paragraph caption, the sentences should be detailed and specific, and the subjects of all sentences MUST be HO.
Please follow the format:\\
\textbf{HO}: ...\\
\textbf{HO's attributes}: ...\\
\textbf{All actions done by HO}: ...\\
\textbf{All statuses of HO}: ...\\
\textbf{All other objects interacted with HO}: ...\\
\textbf{All environments/backgrounds of HO}: ...\\
\textbf{All events related to HO}: ...\\
\textbf{Final object-centric paragraph caption}: The HO is [attributes], [environment]. From ... to ...s, the HO [status], [any action], [any status/attribute/environment changes]... From ... to ...s, the HO [status], [any action], [any status/attribute/environment changes]. The OH's [final status] is ...
\end{tcolorbox}

\begin{figure*}[!ht]
    \centering
    \includegraphics[width=1\linewidth]{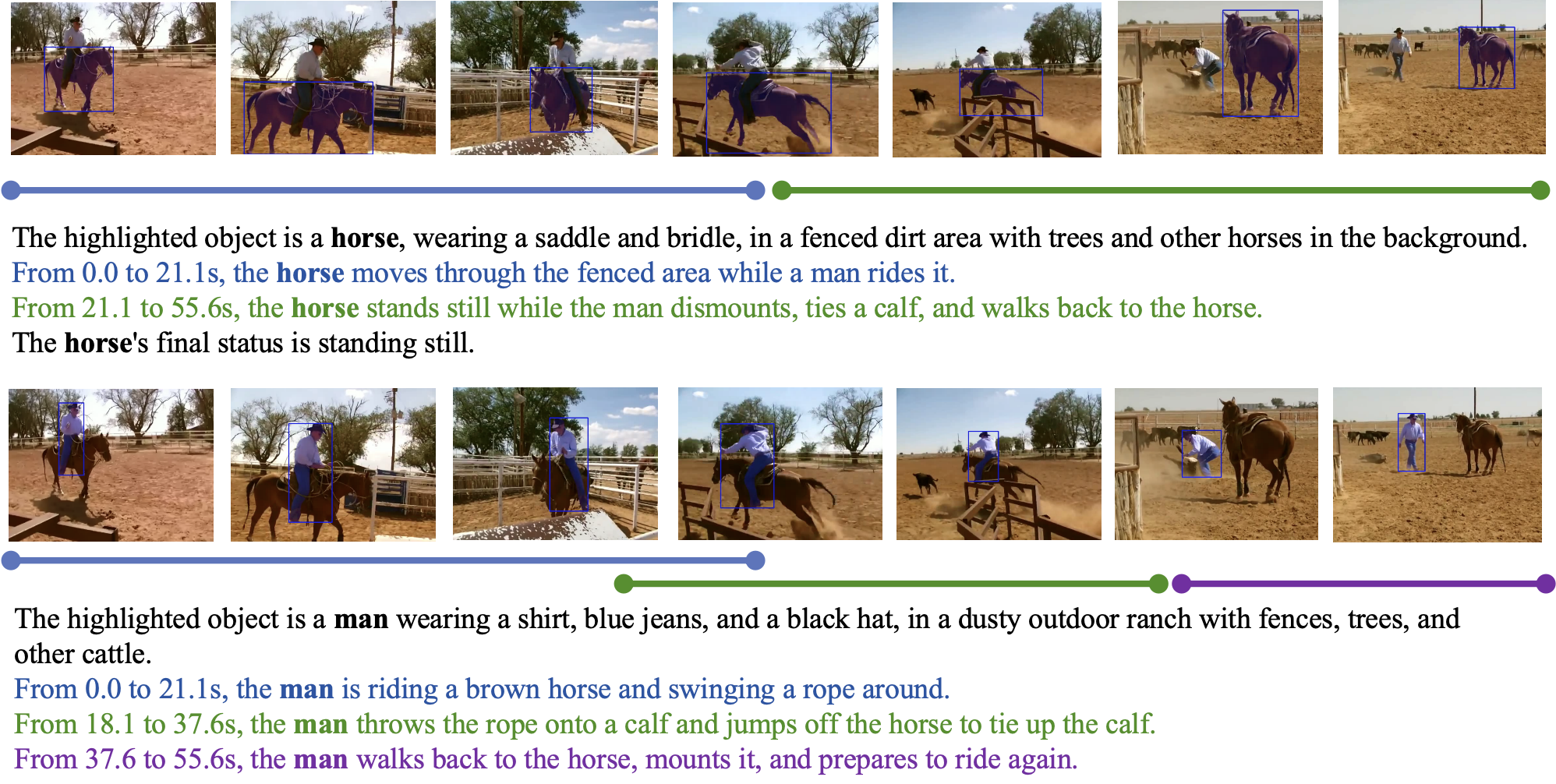}
    \caption{CAT-V can focus on different objects within the same video. The top sequence shows object-centric captioning for a horse, while the bottom sequence demonstrates captioning for the cowboy, each with precise temporal segmentation of their respective actions and states.}
    \label{fig:diff_object}
\end{figure*}

\begin{figure*}[!ht]
    \centering
    \includegraphics[width=0.9\linewidth]{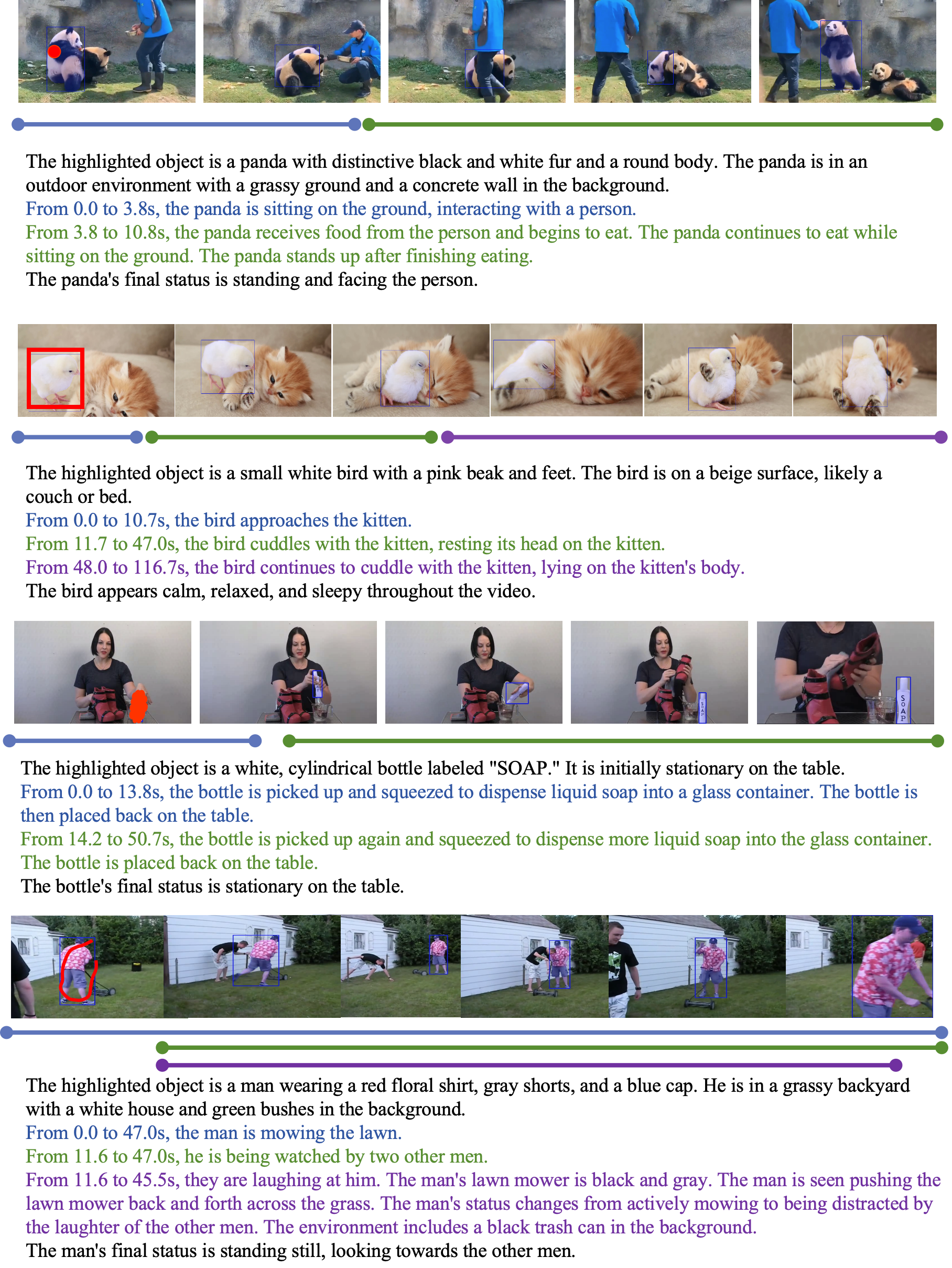}
    \caption{Examples of CAT-V's support for various visual prompting formats. The system effectively handles points, bounding boxes, and irregular regions to identify and track diverse objects including pandas, birds, bottles, and people, demonstrating its flexibility and accuracy in accommodating different user input preferences.}
    \label{fig:diff_vprompt}
\end{figure*}

\begin{figure*}[!ht]
    \centering
    \includegraphics[width=1\linewidth]{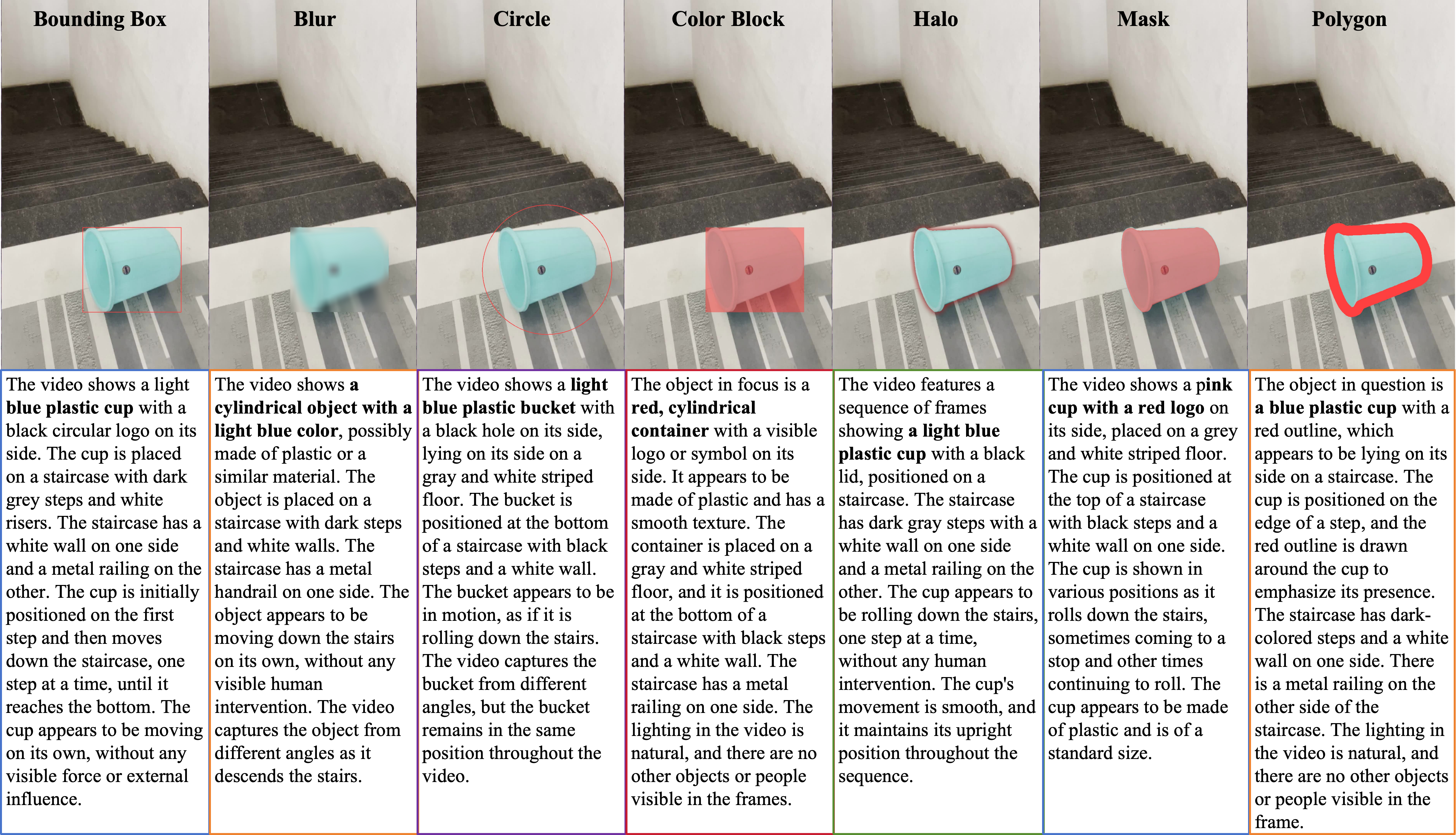}
    \caption{Comparison of different visual prompt styles (Bounding Box, Blur, Circle, Color Block, Halo, Mask, and Polygon) for highlighting a blue plastic cup, demonstrating their effects on object-centric captioning accuracy.}
    \label{fig:diff_vid_prompt}
\end{figure*}

\begin{figure*}[!ht]
    \centering
    \includegraphics[width=\linewidth]{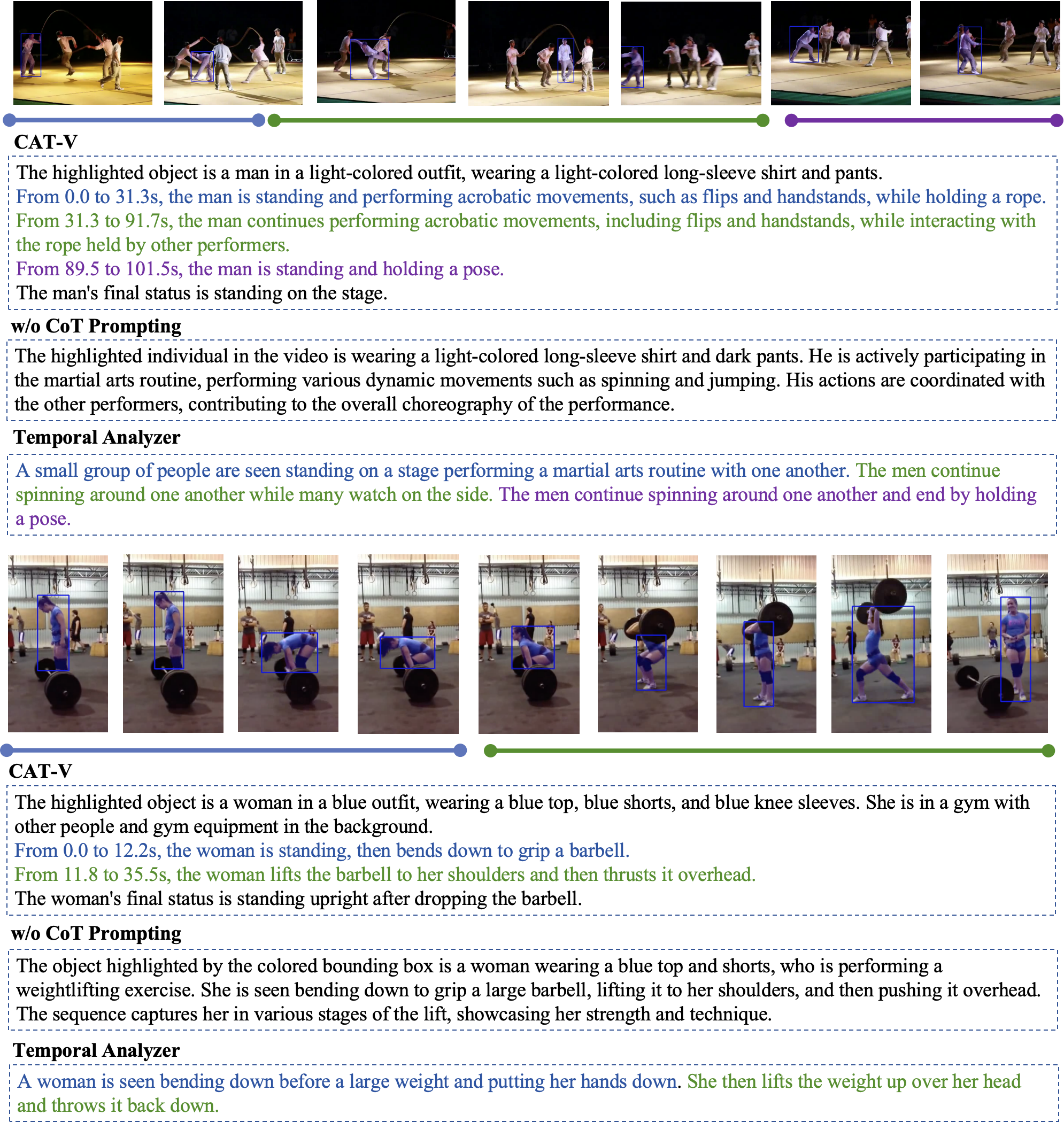}
    \caption{Comparison of object-centric video captioning using CAT-V with CoT prompting, without CoT prompting, and using only the Temporal Analyzer.}
    \label{fig:language_prompt}
\end{figure*}

\begin{figure*}[!ht]
    \centering
    \includegraphics[width=\linewidth]{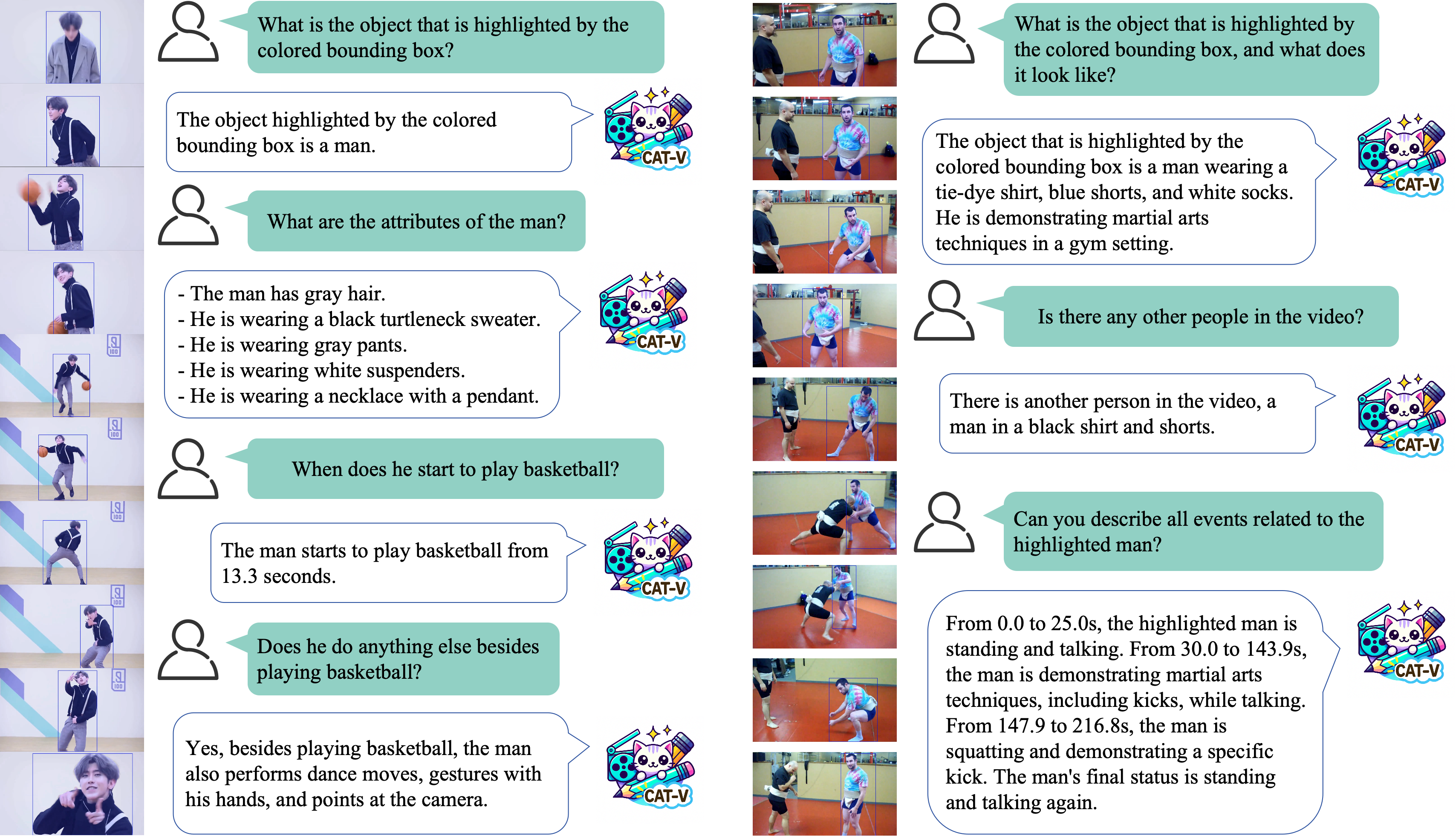}
    \caption{Example of object-centric multi-round chatting with CAT-V, demonstrating the system's ability to maintain reference to the highlighted object while answering specific questions about its attributes and actions.}
    \label{fig:chatting}
\end{figure*}

\section{Experiments}
In this section, we use extensive qualitative experiments to demonstrate the versatility and effectiveness of CAT-V in object-centric video captioning through various visual prompting, highlight styles, Chain-of-Thought prompting, and interactive chatting capabilities.

\subsection{User Visual Prompts}
CAT-V supports versatile user interactions through various visual prompting mechanisms.
As demonstrated in \Cref{fig:diff_object}, users can selectively highlight different objects within the same video for fine-grained captioning.
In this example, the user can choose either the horse or the cowboy to generate object-centric temporal descriptions, with CAT-V accurately tracking and describing the selected entity's actions and state changes throughout the video.
\Cref{fig:diff_vprompt} further illustrates CAT-V's flexibility in accepting different types of visual prompts, including points, bounding boxes, trajectories, and irregular regions.
This adaptability allows users to precisely indicate their object of interest using the most convenient or appropriate prompt type for the particular video content, while CAT-V maintains consistent accuracy in segmentation and captioning regardless of the prompt format.

\subsection{SAM-generated Video Prompts}
\label{sec:injection}
CAT-V leverages SAM 2 to generate masklets of user-selected objects throughout the video and injects these visual cues directly into the video frames as highlighted regions.
These SAM-generated video prompts guide the MLLM to focus on the specific object of interest during captioning.
Figure \ref{fig:diff_vid_prompt} illustrates an experiment comparing different highlight styles for injecting these visual prompts into the video frames.
In this experiment, we bypass both the Temporal Analyzer and CoT Prompting components, directly feeding the prompt-injected video to the MLLM to evaluate the effectiveness of different visual prompt styles.
The results show that bounding boxes and polygons produce the most accurate object-centric descriptions, while other methods like color block and mask tend to alter the object's original appearance, causing the MLLM to generate incorrect descriptions (e.g., identifying a blue cup as "pink" or "red" when color blocks or masks are applied).
Blur and circle methods, while preserving the object's color, provide less precise spatial guidance, sometimes resulting in generic or imprecise descriptions of the object's attributes and movements.

\subsection{Chain-of-Thought Prompts}
As shown in \Cref{fig:language_prompt}, we compare CAT-V's captioning with and without Chain-of-Thought (CoT) prompting. With CoT, the system produces detailed temporal descriptions of highlighted objects, \textit{i.e.}, a performer is doing acrobatics and a woman is lifting weights, specifying precise time intervals and action sequences.
Without CoT, descriptions become generic, lacking temporal precision and detailed object focus.
The Temporal Analyzer provides basic scene descriptions without object-specific details, demonstrating how CoT prompting significantly enhances object-centric video captioning quality.

\subsection{Object-centric Chatting}
CAT-V not only supports fine-grained object-centric video captioning but also enables interactive multi-round chatting focused on specific objects. 
As shown in \Cref{fig:chatting}, users can engage in detailed conversations about the highlighted object, asking follow-up questions to explore its attributes, actions, and temporal behaviors.
This conversational capability allows users to naturally explore different aspects of the object's appearance and behavior in the video through an intuitive dialogue interface.
\section{Related Work}

\subsection{Dense Video Captioning}
The dense video captioning task aims to localize and describe events in a given video by considering the interaction of the object, the spatial location, and the temporal information. The dense video captioning procedure can be divided into three steps: extraction of video features, localization of temporal events, and generation of captions. Previous works \citep{iashin2020better,iashin2020multi,krishna2017dense,wang2018bidirectional,wang2020event,Bi_2021} have performed event localization and caption generation individually. More recent approaches such as PDVC \citep{zhu2022end} and TRACE-Uni~\citep{guo2024trace} jointly estimate event timestamps and captions. PDVC utilizes a DETR-like model \citep{carion2020end}, and TRACD-Uni uses a Large Language Model \citep{jiang2023mistral} as the backbone for end-to-end prediction.

\subsection{Video Object Segmentation}
The video object segmentation task consists of first-frame video object segmentation \citep{pont20172017} and interactive video object segmentation. In this paper, we focus on the interactive video object segmentation task, where user guidance is given as bounding boxes, points, or scribbles. The interactive video object segmentation task has gained a lot of attention recently due to its convenient annotation and intuitive interaction between users and segmentation models. After obtaining user guidance, some works \citep{heo2020interactive,cheng2021modular,delatolas2024learning} design modular approaches to convert user input to a mask in the first video frame and propagate this mask to the remaining video frames sequentially. After the presence of the SAM model, some works \citep{cheng2023tracking,yang2023track,cheng2023segment,rajivc2023segment} propose combining the image-based SAM model with video trackers to enable the video-based segmentation feature. However, in some cases, these hybrid models fail because the video tracker model amplifies and propagates errors caused by the image-based SAM model. Later, \citet{ravi2024sam} proposed a unified segmentation model (SAM 2) that natively supports video object segmentation with memory attention. SAMURAI~\citep{yang2024samurai} further enhances the segmentation capability of SAM 2 by integrating Kalman Filer \citep{kalman1960new} and motion-aware memory into SAM 2.

\subsection{Multimodal Large Language Models}
Vision-Language models \citep{radford2021learning,liu2024visual,hua2024v2xum,xu2023mplug,tang2024avicuna,chen2024far,li2022blip,tong2024cambrian,hua2024finecaption, xie2024characterizing,zhang2024kinmo,Bi_2024} seek multimodal intelligence by jointly processing visual and linguistic information. Inspired by the remarkable success of recent large language models (LLMs)~\citep{touvron2023llama,chiang2023vicuna,hua2023improving}, researchers are now exploring large VLMs that combine pretrained visual encoders and language decoders to tackle complex multimodal tasks. Flamingo~\citep{alayrac2022flamingo} and BLIP-2~\citep{li2023blip} are two of the early works that explore the integration of LLMs into vision-language pre-training. These models are trained as VL foundation models. Beginning with LLaVA~\citep{liu2024visual}, researchers have used LLM-synthesized instruction-following chat data in VQA format for instruction-tuning, achieving significantly improved results~\citep{hua2024finematch,hua2024mmcomposition,yu2023mm, tang2024vidcomposition,bi2024unveilingvisualperceptionlanguage}. Subsequent work has further broadened the capabilities~\citep{hu2023promptcap, hua2024mmcomposition, lin2023videoxum,martin2025wikivideo, yu2024promptfix,lin2024battleagent,bi2023misar}, of multimodal LLMs. However, comparatively little effort has been focused on improving the ability of models to track and describe video content by attending to specific temporal segments and regions.

\section{Conclusion}
We presented CAT-V, a training-free framework for object-centric video captioning that addresses fundamental limitations in existing video understanding approaches.
By integrating SAMURAI's robust object segmentation capabilities, TRACE-Uni's hierarchical temporal analysis, and InternVL-2.5's multimodal understanding, our system enables fine-grained, temporally-aware descriptions of user-selected objects without requiring additional training data.
The use of CoT  guides the model to systematically analyze object attributes, actions, status changes, and interactions, resulting in comprehensive and coherent captions. Our experiments demonstrate CAT-V's versatility in supporting various visual prompt types (points, bounding boxes, and irregular regions) and its effectiveness in maintaining object focus across temporal boundaries.
The system also enables natural conversational interaction about highlighted objects, allowing users to explore specific aspects of video content through intuitive dialogue.
Future work could explore extending CAT-V to handle complex multi-object interactions, incorporating more sophisticated temporal reasoning capabilities, and enhancing its ability to understand causal relationships between objects and events in videos.

\section{Limitations}
Despite CAT-V's capabilities in object-centric video captioning, several limitations remain.
First, CAT-V relies heavily on the segmentation quality of SAMURAI, which may struggle with highly complex scenes, fast motion, or severe occlusions.
When segmentation fails, the subsequent captioning quality degrades significantly.
Second, the framework's temporal accuracy depends on TRACE-Uni's event boundary detection, which can be imprecise for subtle state changes or when multiple events overlap.
Third, while our interactive approach allows flexible object selection, CAT-V currently lacks the ability to handle multiple highlighted objects simultaneously, limiting analysis of object interactions. 

\section*{Acknowledgements}
This work was supported by Sony Group Corporation. We would like to thank Sayaka Nakamura and Jerry Jun Yokono for their insightful discussion.






\bibliography{reference}

\begin{thebibliography}{64}
\providecommand{\natexlab}[1]{#1}
\providecommand{\url}[1]{\texttt{#1}}
\expandafter\ifx\csname urlstyle\endcsname\relax
  \providecommand{\doi}[1]{doi: #1}\else
  \providecommand{\doi}{doi: \begingroup \urlstyle{rm}\Url}\fi

\bibitem[Alayrac et~al.(2022)Alayrac, Donahue, Luc, Miech, Barr, Hasson, Lenc, Mensch, Millican, Reynolds, et~al.]{alayrac2022flamingo}
Jean-Baptiste Alayrac, Jeff Donahue, Pauline Luc, Antoine Miech, Iain Barr, Yana Hasson, Karel Lenc, Arthur Mensch, Katherine Millican, Malcolm Reynolds, et~al.
\newblock Flamingo: a visual language model for few-shot learning.
\newblock \emph{Advances in neural information processing systems}, 35:\penalty0 23716--23736, 2022.

\bibitem[Bi et~al.(2021)Bi, Luo, and Xu]{Bi_2021}
Jing Bi, Jiebo Luo, and Chenliang Xu.
\newblock Procedure planning in instructional videos via contextual modeling and model-based policy learning.
\newblock In \emph{2021 IEEE/CVF International Conference on Computer Vision (ICCV)}, pp.\  15591–15600. IEEE, October 2021.
\newblock \doi{10.1109/iccv48922.2021.01532}.

\bibitem[Bi et~al.(2023)Bi, Nguyen, Vosoughi, and Xu]{bi2023misar}
Jing Bi, Nguyen~Manh Nguyen, Ali Vosoughi, and Chenliang Xu.
\newblock Misar: A multimodal instructional system with augmented reality, 2023.

\bibitem[Bi et~al.(2024{\natexlab{a}})Bi, Guo, Tang, Wen, Liu, and Xu]{bi2024unveilingvisualperceptionlanguage}
Jing Bi, Junjia Guo, Yunlong Tang, Lianggong~Bruce Wen, Zhang Liu, and Chenliang Xu.
\newblock Unveiling visual perception in language models: An attention head analysis approach, 2024{\natexlab{a}}.

\bibitem[Bi et~al.(2024{\natexlab{b}})Bi, Tang, Song, Vosoughi, Nguyen, and Xu]{Bi_2024}
Jing Bi, Yunlong Tang, Luchuan Song, Ali Vosoughi, Nguyen Nguyen, and Chenliang Xu.
\newblock Eagle: Egocentric aggregated language-video engine.
\newblock In \emph{Proceedings of the 32nd ACM International Conference on Multimedia}, MM ’24, pp.\  1682–1691. ACM, October 2024{\natexlab{b}}.
\newblock \doi{10.1145/3664647.3681618}.

\bibitem[Carion et~al.(2020)Carion, Massa, Synnaeve, Usunier, Kirillov, and Zagoruyko]{carion2020end}
Nicolas Carion, Francisco Massa, Gabriel Synnaeve, Nicolas Usunier, Alexander Kirillov, and Sergey Zagoruyko.
\newblock End-to-end object detection with transformers.
\newblock In \emph{European conference on computer vision}, pp.\  213--229. Springer, 2020.

\bibitem[Chen et~al.(2024{\natexlab{a}})Chen, Wang, Cao, Liu, Gao, Cui, Zhu, Ye, Tian, Liu, et~al.]{chen2024internvl2.5}
Zhe Chen, Weiyun Wang, Yue Cao, Yangzhou Liu, Zhangwei Gao, Erfei Cui, Jinguo Zhu, Shenglong Ye, Hao Tian, Zhaoyang Liu, et~al.
\newblock Expanding performance boundaries of open-source multimodal models with model, data, and test-time scaling.
\newblock \emph{arXiv preprint arXiv:2412.05271}, 2024{\natexlab{a}}.

\bibitem[Chen et~al.(2024{\natexlab{b}})Chen, Wang, Tian, Ye, Gao, Cui, Tong, Hu, Luo, Ma, et~al.]{chen2024far}
Zhe Chen, Weiyun Wang, Hao Tian, Shenglong Ye, Zhangwei Gao, Erfei Cui, Wenwen Tong, Kongzhi Hu, Jiapeng Luo, Zheng Ma, et~al.
\newblock How far are we to gpt-4v? closing the gap to commercial multimodal models with open-source suites.
\newblock \emph{arXiv preprint arXiv:2404.16821}, 2024{\natexlab{b}}.

\bibitem[Chen et~al.(2024{\natexlab{c}})Chen, Wu, Wang, Su, Chen, Xing, Zhong, Zhang, Zhu, Lu, et~al.]{chen2024internvl}
Zhe Chen, Jiannan Wu, Wenhai Wang, Weijie Su, Guo Chen, Sen Xing, Muyan Zhong, Qinglong Zhang, Xizhou Zhu, Lewei Lu, et~al.
\newblock Internvl: Scaling up vision foundation models and aligning for generic visual-linguistic tasks.
\newblock In \emph{Proceedings of the IEEE/CVF Conference on Computer Vision and Pattern Recognition}, pp.\  24185--24198, 2024{\natexlab{c}}.

\bibitem[Cheng et~al.(2021)Cheng, Tai, and Tang]{cheng2021modular}
Ho~Kei Cheng, Yu-Wing Tai, and Chi-Keung Tang.
\newblock Modular interactive video object segmentation: Interaction-to-mask, propagation and difference-aware fusion.
\newblock In \emph{Proceedings of the IEEE/CVF Conference on Computer Vision and Pattern Recognition}, pp.\  5559--5568, 2021.

\bibitem[Cheng et~al.(2023{\natexlab{a}})Cheng, Oh, Price, Schwing, and Lee]{cheng2023tracking}
Ho~Kei Cheng, Seoung~Wug Oh, Brian Price, Alexander Schwing, and Joon-Young Lee.
\newblock Tracking anything with decoupled video segmentation.
\newblock In \emph{Proceedings of the IEEE/CVF International Conference on Computer Vision}, pp.\  1316--1326, 2023{\natexlab{a}}.

\bibitem[Cheng et~al.(2023{\natexlab{b}})Cheng, Li, Xu, Li, Yang, Wang, and Yang]{cheng2023segment}
Yangming Cheng, Liulei Li, Yuanyou Xu, Xiaodi Li, Zongxin Yang, Wenguan Wang, and Yi~Yang.
\newblock Segment and track anything.
\newblock \emph{arXiv preprint arXiv:2305.06558}, 2023{\natexlab{b}}.

\bibitem[Chiang et~al.(2023)Chiang, Li, Lin, Sheng, Wu, Zhang, Zheng, Zhuang, Zhuang, Gonzalez, et~al.]{chiang2023vicuna}
Wei-Lin Chiang, Zhuohan Li, Zi~Lin, Ying Sheng, Zhanghao Wu, Hao Zhang, Lianmin Zheng, Siyuan Zhuang, Yonghao Zhuang, Joseph~E Gonzalez, et~al.
\newblock Vicuna: An open-source chatbot impressing gpt-4 with 90\%* chatgpt quality, march 2023.
\newblock \emph{URL https://lmsys. org/blog/2023-03-30-vicuna}, 3\penalty0 (5), 2023.

\bibitem[Delatolas et~al.(2024)Delatolas, Kalogeiton, and Papadopoulos]{delatolas2024learning}
Thanos Delatolas, Vicky Kalogeiton, and Dim~P Papadopoulos.
\newblock Learning the what and how of annotation in video object segmentation.
\newblock In \emph{Proceedings of the IEEE/CVF Winter Conference on Applications of Computer Vision}, pp.\  6951--6961, 2024.

\bibitem[Guo et~al.(2024)Guo, Liu, Li, Tang, Liu, and Chen]{guo2024trace}
Yongxin Guo, Jingyu Liu, Mingda Li, Xiaoying Tang, Qingbin Liu, and Xi~Chen.
\newblock Trace: Temporal grounding video llm via causal event modeling.
\newblock \emph{arXiv preprint arXiv:2410.05643}, 2024.

\bibitem[Heo et~al.(2020)Heo, Jun~Koh, and Kim]{heo2020interactive}
Yuk Heo, Yeong Jun~Koh, and Chang-Su Kim.
\newblock Interactive video object segmentation using global and local transfer modules.
\newblock In \emph{Computer Vision--ECCV 2020: 16th European Conference, Glasgow, UK, August 23--28, 2020, Proceedings, Part XVII 16}, pp.\  297--313. Springer, 2020.

\bibitem[Hu et~al.(2023)Hu, Hua, Yang, Shi, Smith, and Luo]{hu2023promptcap}
Yushi Hu, Hang Hua, Zhengyuan Yang, Weijia Shi, Noah~A Smith, and Jiebo Luo.
\newblock Promptcap: Prompt-guided image captioning for vqa with gpt-3.
\newblock In \emph{Proceedings of the IEEE/CVF International Conference on Computer Vision}, pp.\  2963--2975, 2023.

\bibitem[Hua et~al.(2023)Hua, Li, Dou, Xu, and Luo]{hua2023improving}
Hang Hua, Xingjian Li, Dejing Dou, Cheng-Zhong Xu, and Jiebo Luo.
\newblock Improving pretrained language model fine-tuning with noise stability regularization.
\newblock \emph{IEEE Transactions on Neural Networks and Learning Systems}, 2023.

\bibitem[Hua et~al.(2024{\natexlab{a}})Hua, Liu, Zhang, Shi, Zhang, Wang, Zhang, and Luo]{hua2024finecaption}
Hang Hua, Qing Liu, Lingzhi Zhang, Jing Shi, Zhifei Zhang, Yilin Wang, Jianming Zhang, and Jiebo Luo.
\newblock Finecaption: Compositional image captioning focusing on wherever you want at any granularity.
\newblock \emph{arXiv preprint arXiv:2411.15411}, 2024{\natexlab{a}}.

\bibitem[Hua et~al.(2024{\natexlab{b}})Hua, Shi, Kafle, Jenni, Zhang, Collomosse, Cohen, and Luo]{hua2024finematch}
Hang Hua, Jing Shi, Kushal Kafle, Simon Jenni, Daoan Zhang, John Collomosse, Scott Cohen, and Jiebo Luo.
\newblock Finematch: Aspect-based fine-grained image and text mismatch detection and correction.
\newblock \emph{arXiv preprint arXiv:2404.14715}, 2024{\natexlab{b}}.

\bibitem[Hua et~al.(2024{\natexlab{c}})Hua, Tang, Xu, and Luo]{hua2024v2xum}
Hang Hua, Yunlong Tang, Chenliang Xu, and Jiebo Luo.
\newblock V2xum-llm: Cross-modal video summarization with temporal prompt instruction tuning.
\newblock \emph{arXiv preprint arXiv:2404.12353}, 2024{\natexlab{c}}.

\bibitem[Hua et~al.(2024{\natexlab{d}})Hua, Tang, Zeng, Cao, Yang, He, Xu, and Luo]{hua2024mmcomposition}
Hang Hua, Yunlong Tang, Ziyun Zeng, Liangliang Cao, Zhengyuan Yang, Hangfeng He, Chenliang Xu, and Jiebo Luo.
\newblock Mmcomposition: Revisiting the compositionality of pre-trained vision-language models.
\newblock \emph{arXiv preprint arXiv:2410.09733}, 2024{\natexlab{d}}.

\bibitem[Huang et~al.(2024{\natexlab{a}})Huang, Wang, Chen, Song, and Zhu]{huang2024vtimellm}
Bin Huang, Xin Wang, Hong Chen, Zihan Song, and Wenwu Zhu.
\newblock Vtimellm: Empower llm to grasp video moments.
\newblock In \emph{Proceedings of the IEEE/CVF Conference on Computer Vision and Pattern Recognition}, pp.\  14271--14280, 2024{\natexlab{a}}.

\bibitem[Huang et~al.(2023)Huang, Tian, Kumar, and Xu]{Huang_2023_CVPR}
Chao Huang, Yapeng Tian, Anurag Kumar, and Chenliang Xu.
\newblock Egocentric audio-visual object localization.
\newblock In \emph{Proceedings of the IEEE/CVF Conference on Computer Vision and Pattern Recognition (CVPR)}, pp.\  22910--22921, June 2023.

\bibitem[Huang et~al.(2024{\natexlab{b}})Huang, Wang, Tang, Zhang, Hu, Lu, Wang, and Liu]{huang2024segment}
Xiaoke Huang, Jianfeng Wang, Yansong Tang, Zheng Zhang, Han Hu, Jiwen Lu, Lijuan Wang, and Zicheng Liu.
\newblock Segment and caption anything.
\newblock In \emph{Proceedings of the IEEE/CVF conference on computer vision and pattern recognition}, pp.\  13405--13417, 2024{\natexlab{b}}.

\bibitem[Iashin \& Rahtu(2020{\natexlab{a}})Iashin and Rahtu]{iashin2020better}
Vladimir Iashin and Esa Rahtu.
\newblock A better use of audio-visual cues: Dense video captioning with bi-modal transformer.
\newblock \emph{arXiv preprint arXiv:2005.08271}, 2020{\natexlab{a}}.

\bibitem[Iashin \& Rahtu(2020{\natexlab{b}})Iashin and Rahtu]{iashin2020multi}
Vladimir Iashin and Esa Rahtu.
\newblock Multi-modal dense video captioning.
\newblock In \emph{Proceedings of the IEEE/CVF conference on computer vision and pattern recognition workshops}, pp.\  958--959, 2020{\natexlab{b}}.

\bibitem[Jiang et~al.(2023)Jiang, Sablayrolles, Mensch, Bamford, Chaplot, Casas, Bressand, Lengyel, Lample, Saulnier, et~al.]{jiang2023mistral}
Albert~Q Jiang, Alexandre Sablayrolles, Arthur Mensch, Chris Bamford, Devendra~Singh Chaplot, Diego de~las Casas, Florian Bressand, Gianna Lengyel, Guillaume Lample, Lucile Saulnier, et~al.
\newblock Mistral 7b.
\newblock \emph{arXiv preprint arXiv:2310.06825}, 2023.

\bibitem[Kalman(1960)]{kalman1960new}
Rudolph~Emil Kalman.
\newblock A new approach to linear filtering and prediction problems.
\newblock 1960.

\bibitem[Krishna et~al.(2017)Krishna, Hata, Ren, Fei-Fei, and Carlos~Niebles]{krishna2017dense}
Ranjay Krishna, Kenji Hata, Frederic Ren, Li~Fei-Fei, and Juan Carlos~Niebles.
\newblock Dense-captioning events in videos.
\newblock In \emph{Proceedings of the IEEE international conference on computer vision}, pp.\  706--715, 2017.

\bibitem[Li et~al.(2022)Li, Li, Xiong, and Hoi]{li2022blip}
Junnan Li, Dongxu Li, Caiming Xiong, and Steven Hoi.
\newblock Blip: Bootstrapping language-image pre-training for unified vision-language understanding and generation.
\newblock In \emph{International conference on machine learning}, pp.\  12888--12900. PMLR, 2022.

\bibitem[Li et~al.(2023{\natexlab{a}})Li, Li, Savarese, and Hoi]{li2023blip}
Junnan Li, Dongxu Li, Silvio Savarese, and Steven Hoi.
\newblock Blip-2: Bootstrapping language-image pre-training with frozen image encoders and large language models.
\newblock In \emph{International conference on machine learning}, pp.\  19730--19742. PMLR, 2023{\natexlab{a}}.

\bibitem[Li et~al.(2023{\natexlab{b}})Li, He, Wang, Li, Wang, Luo, Wang, Wang, and Qiao]{li2023videochat}
KunChang Li, Yinan He, Yi~Wang, Yizhuo Li, Wenhai Wang, Ping Luo, Yali Wang, Limin Wang, and Yu~Qiao.
\newblock Videochat: Chat-centric video understanding.
\newblock \emph{arXiv preprint arXiv:2305.06355}, 2023{\natexlab{b}}.

\bibitem[Lin et~al.(2023)Lin, Hua, Chen, Li, Hsiao, Ho, and Luo]{lin2023videoxum}
Jingyang Lin, Hang Hua, Ming Chen, Yikang Li, Jenhao Hsiao, Chiuman Ho, and Jiebo Luo.
\newblock Videoxum: Cross-modal visual and textural summarization of videos.
\newblock \emph{IEEE Transactions on Multimedia}, 2023.

\bibitem[Lin et~al.(2024)Lin, Hua, Li, Chang, Fan, Ji, Hua, Jin, Luo, and Zhang]{lin2024battleagent}
Shuhang Lin, Wenyue Hua, Lingyao Li, Che-Jui Chang, Lizhou Fan, Jianchao Ji, Hang Hua, Mingyu Jin, Jiebo Luo, and Yongfeng Zhang.
\newblock Battleagent: Multi-modal dynamic emulation on historical battles to complement historical analysis.
\newblock \emph{arXiv preprint arXiv:2404.15532}, 2024.

\bibitem[Liu et~al.(2024)Liu, Li, Wu, and Lee]{liu2024visual}
Haotian Liu, Chunyuan Li, Qingyang Wu, and Yong~Jae Lee.
\newblock Visual instruction tuning.
\newblock \emph{Advances in neural information processing systems}, 36, 2024.

\bibitem[Maaz et~al.(2023)Maaz, Rasheed, Khan, and Khan]{maaz2023videochatgpt}
Muhammad Maaz, Hanoona Rasheed, Salman Khan, and Fahad~Shahbaz Khan.
\newblock Video-chatgpt: Towards detailed video understanding via large vision and language models.
\newblock \emph{arXiv preprint arXiv:2306.05424}, 2023.

\bibitem[Martin et~al.(2025)Martin, Kriz, Walden, Sanders, Recknor, Yang, Ferraro, and Van~Durme]{martin2025wikivideo}
Alexander Martin, Reno Kriz, William~Gantt Walden, Kate Sanders, Hannah Recknor, Eugene Yang, Francis Ferraro, and Benjamin Van~Durme.
\newblock Wikivideo: Article generation from multiple videos.
\newblock \emph{arXiv preprint arXiv:2504.00939}, 2025.

\bibitem[Pont-Tuset et~al.(2017)Pont-Tuset, Perazzi, Caelles, Arbel{\'a}ez, Sorkine-Hornung, and Van~Gool]{pont20172017}
Jordi Pont-Tuset, Federico Perazzi, Sergi Caelles, Pablo Arbel{\'a}ez, Alex Sorkine-Hornung, and Luc Van~Gool.
\newblock The 2017 davis challenge on video object segmentation.
\newblock \emph{arXiv preprint arXiv:1704.00675}, 2017.

\bibitem[Radford et~al.(2021)Radford, Kim, Hallacy, Ramesh, Goh, Agarwal, Sastry, Askell, Mishkin, Clark, et~al.]{radford2021learning}
Alec Radford, Jong~Wook Kim, Chris Hallacy, Aditya Ramesh, Gabriel Goh, Sandhini Agarwal, Girish Sastry, Amanda Askell, Pamela Mishkin, Jack Clark, et~al.
\newblock Learning transferable visual models from natural language supervision.
\newblock In \emph{International conference on machine learning}, pp.\  8748--8763. PMLR, 2021.

\bibitem[Raji{\v{c}} et~al.(2023)Raji{\v{c}}, Ke, Tai, Tang, Danelljan, and Yu]{rajivc2023segment}
Frano Raji{\v{c}}, Lei Ke, Yu-Wing Tai, Chi-Keung Tang, Martin Danelljan, and Fisher Yu.
\newblock Segment anything meets point tracking.
\newblock \emph{arXiv preprint arXiv:2307.01197}, 2023.

\bibitem[Ravi et~al.(2024)Ravi, Gabeur, Hu, Hu, Ryali, Ma, Khedr, R{\"a}dle, Rolland, Gustafson, et~al.]{ravi2024sam}
Nikhila Ravi, Valentin Gabeur, Yuan-Ting Hu, Ronghang Hu, Chaitanya Ryali, Tengyu Ma, Haitham Khedr, Roman R{\"a}dle, Chloe Rolland, Laura Gustafson, et~al.
\newblock Sam 2: Segment anything in images and videos.
\newblock \emph{arXiv preprint arXiv:2408.00714}, 2024.

\bibitem[Tang et~al.(2023)Tang, Bi, Xu, Song, Liang, Wang, Zhang, An, Lin, Zhu, et~al.]{tang2023video}
Yunlong Tang, Jing Bi, Siting Xu, Luchuan Song, Susan Liang, Teng Wang, Daoan Zhang, Jie An, Jingyang Lin, Rongyi Zhu, et~al.
\newblock Video understanding with large language models: A survey.
\newblock \emph{arXiv preprint arXiv:2312.17432}, 2023.

\bibitem[Tang et~al.(2024{\natexlab{a}})Tang, Guo, Hua, Liang, Feng, Li, Mao, Huang, Bi, Zhang, et~al.]{tang2024vidcomposition}
Yunlong Tang, Junjia Guo, Hang Hua, Susan Liang, Mingqian Feng, Xinyang Li, Rui Mao, Chao Huang, Jing Bi, Zeliang Zhang, et~al.
\newblock Vidcomposition: Can mllms analyze compositions in compiled videos?
\newblock \emph{arXiv preprint arXiv:2411.10979}, 2024{\natexlab{a}}.

\bibitem[Tang et~al.(2024{\natexlab{b}})Tang, Shimada, Bi, Feng, Hua, and Xu]{tang2024avicuna}
Yunlong Tang, Daiki Shimada, Jing Bi, Mingqian Feng, Hang Hua, and Chenliang Xu.
\newblock Empowering llms with pseudo-untrimmed videos for audio-visual temporal understanding.
\newblock \emph{arXiv preprint arXiv:2403.16276}, 2024{\natexlab{b}}.

\bibitem[Tong et~al.(2024)Tong, Brown, Wu, Woo, Middepogu, Akula, Yang, Yang, Iyer, Pan, et~al.]{tong2024cambrian}
Shengbang Tong, Ellis Brown, Penghao Wu, Sanghyun Woo, Manoj Middepogu, Sai~Charitha Akula, Jihan Yang, Shusheng Yang, Adithya Iyer, Xichen Pan, et~al.
\newblock Cambrian-1: A fully open, vision-centric exploration of multimodal llms.
\newblock \emph{arXiv preprint arXiv:2406.16860}, 2024.

\bibitem[Touvron et~al.(2023)Touvron, Lavril, Izacard, Martinet, Lachaux, Lacroix, Rozi{\`e}re, Goyal, Hambro, Azhar, et~al.]{touvron2023llama}
Hugo Touvron, Thibaut Lavril, Gautier Izacard, Xavier Martinet, Marie-Anne Lachaux, Timoth{\'e}e Lacroix, Baptiste Rozi{\`e}re, Naman Goyal, Eric Hambro, Faisal Azhar, et~al.
\newblock Llama: Open and efficient foundation language models.
\newblock \emph{arXiv preprint arXiv:2302.13971}, 2023.

\bibitem[Wang et~al.(2018)Wang, Jiang, Ma, Liu, and Xu]{wang2018bidirectional}
Jingwen Wang, Wenhao Jiang, Lin Ma, Wei Liu, and Yong Xu.
\newblock Bidirectional attentive fusion with context gating for dense video captioning.
\newblock In \emph{Proceedings of the IEEE conference on computer vision and pattern recognition}, pp.\  7190--7198, 2018.

\bibitem[Wang et~al.(2020)Wang, Zheng, Yu, Tian, and Hu]{wang2020event}
Teng Wang, Huicheng Zheng, Mingjing Yu, Qian Tian, and Haifeng Hu.
\newblock Event-centric hierarchical representation for dense video captioning.
\newblock \emph{IEEE Transactions on Circuits and Systems for Video Technology}, 31\penalty0 (5):\penalty0 1890--1900, 2020.

\bibitem[Wang et~al.(2021)Wang, Zhang, Lu, Zheng, Cheng, and Luo]{wang2021pdvc}
Teng Wang, Ruimao Zhang, Zhichao Lu, Feng Zheng, Ran Cheng, and Ping Luo.
\newblock End-to-end dense video captioning with parallel decoding.
\newblock In \emph{Proceedings of the IEEE/CVF international conference on computer vision}, pp.\  6847--6857, 2021.

\bibitem[Wang et~al.(2023)Wang, Zhang, Fei, Zheng, Tang, Li, Gao, and Zhao]{wang2023caption}
Teng Wang, Jinrui Zhang, Junjie Fei, Hao Zheng, Yunlong Tang, Zhe Li, Mingqi Gao, and Shanshan Zhao.
\newblock Caption anything: Interactive image description with diverse multimodal controls.
\newblock \emph{arXiv preprint arXiv:2305.02677}, 2023.

\bibitem[Xie et~al.(2024)Xie, Deng, Liu, Lou, Xu, and Li]{xie2024characterizing}
Zidian Xie, Shijian Deng, Pinxin Liu, Xubin Lou, Chenliang Xu, and Dongmei Li.
\newblock Characterizing anti-vaping posts for effective communication on instagram using multimodal deep learning.
\newblock \emph{Nicotine and Tobacco Research}, 26\penalty0 (Supplement\_1):\penalty0 S43--S48, 2024.

\bibitem[Xu et~al.(2023)Xu, Ye, Yan, Shi, Ye, Xu, Li, Bi, Qian, Wang, et~al.]{xu2023mplug}
Haiyang Xu, Qinghao Ye, Ming Yan, Yaya Shi, Jiabo Ye, Yuanhong Xu, Chenliang Li, Bin Bi, Qi~Qian, Wei Wang, et~al.
\newblock mplug-2: A modularized multi-modal foundation model across text, image and video.
\newblock In \emph{International Conference on Machine Learning}, pp.\  38728--38748. PMLR, 2023.

\bibitem[Yang et~al.(2023{\natexlab{a}})Yang, Nagrani, Seo, Miech, Pont-Tuset, Laptev, Sivic, and Schmid]{yang2023vid2seq}
Antoine Yang, Arsha Nagrani, Paul~Hongsuck Seo, Antoine Miech, Jordi Pont-Tuset, Ivan Laptev, Josef Sivic, and Cordelia Schmid.
\newblock Vid2seq: Large-scale pretraining of a visual language model for dense video captioning.
\newblock In \emph{Proceedings of the IEEE/CVF Conference on Computer Vision and Pattern Recognition}, pp.\  10714--10726, 2023{\natexlab{a}}.

\bibitem[Yang et~al.(2024)Yang, Huang, Chai, Jiang, and Hwang]{yang2024samurai}
Cheng-Yen Yang, Hsiang-Wei Huang, Wenhao Chai, Zhongyu Jiang, and Jenq-Neng Hwang.
\newblock Samurai: Adapting segment anything model for zero-shot visual tracking with motion-aware memory.
\newblock \emph{arXiv preprint arXiv:2411.11922}, 2024.

\bibitem[Yang et~al.(2023{\natexlab{b}})Yang, Gao, Li, Gao, Wang, and Zheng]{yang2023track}
Jinyu Yang, Mingqi Gao, Zhe Li, Shang Gao, Fangjing Wang, and Feng Zheng.
\newblock Track anything: Segment anything meets videos.
\newblock \emph{arXiv preprint arXiv:2304.11968}, 2023{\natexlab{b}}.

\bibitem[Yu et~al.(2023)Yu, Yang, Li, Wang, Lin, Liu, Wang, and Wang]{yu2023mm}
Weihao Yu, Zhengyuan Yang, Linjie Li, Jianfeng Wang, Kevin Lin, Zicheng Liu, Xinchao Wang, and Lijuan Wang.
\newblock Mm-vet: Evaluating large multimodal models for integrated capabilities.
\newblock \emph{arXiv preprint arXiv:2308.02490}, 2023.

\bibitem[Yu et~al.(2024)Yu, Zeng, Hua, Fu, and Luo]{yu2024promptfix}
Yongsheng Yu, Ziyun Zeng, Hang Hua, Jianlong Fu, and Jiebo Luo.
\newblock Promptfix: You prompt and we fix the photo.
\newblock \emph{arXiv preprint arXiv:2405.16785}, 2024.

\bibitem[Yuan et~al.(2025)Yuan, Li, Zhang, Huang, Xu, Ji, Tong, Qi, Feng, and Yang]{yuan2025sa2va}
Haobo Yuan, Xiangtai Li, Tao Zhang, Zilong Huang, Shilin Xu, Shunping Ji, Yunhai Tong, Lu~Qi, Jiashi Feng, and Ming-Hsuan Yang.
\newblock Sa2va: Marrying sam2 with llava for dense grounded understanding of images and videos.
\newblock \emph{arXiv preprint arXiv:2501.04001}, 2025.

\bibitem[Zhang et~al.(2025)Zhang, Li, Cheng, Hu, Yuan, Chen, Leng, Jiang, Zhang, Li, et~al.]{zhang2025videollama3}
Boqiang Zhang, Kehan Li, Zesen Cheng, Zhiqiang Hu, Yuqian Yuan, Guanzheng Chen, Sicong Leng, Yuming Jiang, Hang Zhang, Xin Li, et~al.
\newblock Videollama 3: Frontier multimodal foundation models for image and video understanding.
\newblock \emph{arXiv preprint arXiv:2501.13106}, 2025.

\bibitem[Zhang et~al.(2024)Zhang, Liu, Kim, Garrido, and Chaudhuri]{zhang2024kinmo}
Pengfei Zhang, Pinxin Liu, Hyeongwoo Kim, Pablo Garrido, and Bindita Chaudhuri.
\newblock Kinmo: Kinematic-aware human motion understanding and generation.
\newblock \emph{arXiv preprint arXiv:2411.15472}, 2024.

\bibitem[Zhang et~al.(2021)Zhang, Liang, Yang, Liu, Wu, Shan, and Chen]{zhang2021unicon}
Yuanhang Zhang, Susan Liang, Shuang Yang, Xiao Liu, Zhongqin Wu, Shiguang Shan, and Xilin Chen.
\newblock Unicon: Unified context network for robust active speaker detection.
\newblock In \emph{Proceedings of the 29th ACM international conference on multimedia}, pp.\  3964--3972, 2021.

\bibitem[Zhou et~al.(2018)Zhou, Xu, and Corso]{zhou2018towards}
Luowei Zhou, Chenliang Xu, and Jason Corso.
\newblock Towards automatic learning of procedures from web instructional videos.
\newblock In \emph{Proceedings of the AAAI conference on artificial intelligence}, volume~32, 2018.

\bibitem[Zhu et~al.(2022)Zhu, Pang, Thapliyal, Wang, and Soricut]{zhu2022end}
Wanrong Zhu, Bo~Pang, Ashish~V Thapliyal, William~Yang Wang, and Radu Soricut.
\newblock End-to-end dense video captioning as sequence generation.
\newblock \emph{arXiv preprint arXiv:2204.08121}, 2022.

\end{thebibliography}
\bibliographystyle{reference}

\end{document}